%% file: main.tex
\documentclass[10pt,twocolumn,letterpaper]{article}

\usepackage[pagenumbers]{cvpr} % To force page numbers, e.g. for an arXiv version

\input{preamble}

\definecolor{cvprblue}{rgb}{0.21,0.49,0.74}
\usepackage[pagebackref,breaklinks,colorlinks,allcolors=cvprblue]{hyperref}

\def\paperID{15371} % *** Enter the Paper ID here
\def\confName{CVPR}
\def\confYear{2025}

\title{Masked Scene Modeling:\\Narrowing the Gap Between Supervised and Self-Supervised Learning\\in 3D Scene Understanding}

\author{
Pedro Hermosilla\\
TU Wien
\and
Christian Stippel\\
TU Wien
\and
Leon Sick\\
Ulm University
}

\begin{document}

\twocolumn[{
\renewcommand\twocolumn[1][]{#1}
\maketitle
\includegraphics[width=\textwidth]{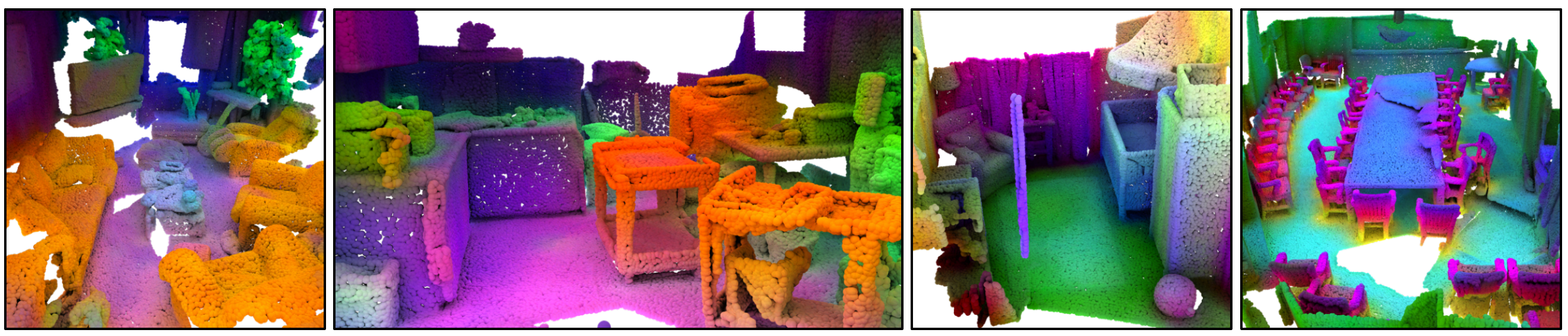}
\captionof{figure}{\textbf{Self-Supervised Feature Visualization using PCA.} 
We reduce the point features obtained with our self-supervised model to three dimensions using PCA and visualize them as colors. 
Features learned by our model are semantic-aware, which is visible from the color separation: Similar objects result in similar features, such as the sofas in the first figure or the chairs in the last one, while different objects result in different features, such as the counter and the tables in the second image or the crib and the curtains in the third one.}
\label{fig:teaser}
\vspace{.5cm}
}]

\input{0_abstract}    
\input{1_intro}
\input{2_related_work}
\input{3_metrics}
\input{4_methods}
\input{5_comp_sota}
\input{6_conclusions}

\newpage
\input{X_suppl}

\newpage
{
    \small
    \bibliographystyle{ieeenat_fullname}
    \bibliography{main}
}

% WARNING: do not forget to delete the supplementary pages from your submission 
% \input{sec/X_suppl}

\end{document}

%% file: preamble.tex
\usepackage[dvipsnames]{xcolor}

\usepackage{multirow}
\usepackage{amssymb}
\usepackage{pifont}% http://ctan.org/pkg/pifont
\usepackage{subcaption}
\usepackage{color}
\usepackage{colortbl}
\usepackage{graphicx}
\usepackage{acronym}
\usepackage{wrapfig}

\usepackage[page,header]{appendix}
\usepackage{titletoc}

\definecolor{CellBck}{gray}{0.92}
\definecolor{SupColor}{RGB}{100, 100, 100}

\newcommand{\cbg}{\cellcolor{CellBck}}

\acrodef{NN}[NN]{Nearest Neighbor}
\acrodef{mIoU}[mIoU]{mean Intersection over Union}
\acrodef{IoU}[IoU]{Intersection over Union}
\acrodef{MHA}[MHA]{Multi-Head Attention}
\acrodef{mAP}[mAP]{mean Average Precission}
\acrodef{MLP}[MLP]{Multi Layer Perceptron}
\acrodef{MIM}[MIM]{Masked Image Modelling}
\acrodef{EMA}[EMA]{Exponential Moving Average}

%% file: 0_abstract.tex
\begin{abstract}
Self-supervised learning has transformed 2D computer vision by enabling models trained on large, unannotated datasets to provide versatile off-the-shelf features that perform similarly to models trained with labels. 
However, in 3D scene understanding, self-supervised methods are typically only used as a weight initialization step for task-specific fine-tuning, limiting their utility for general-purpose feature extraction.  
This paper addresses this shortcoming by proposing a robust evaluation protocol specifically designed to assess the quality of self-supervised features for 3D scene understanding. 
Our protocol uses multi-resolution feature sampling of hierarchical models to create rich point-level representations that capture the semantic capabilities of the model and, hence, are suitable for evaluation with linear probing and nearest-neighbor methods.
Furthermore, we introduce the first self-supervised model that performs similarly to supervised models when only off-the-shelf features are used in a linear probing setup.
In particular, our model is trained natively in 3D with a novel self-supervised approach based on a Masked Scene Modeling objective, which reconstructs deep features of masked patches in a bottom-up manner and is specifically tailored to hierarchical 3D models. 
Our experiments not only demonstrate that our method achieves competitive performance to supervised models, but also surpasses existing self-supervised approaches by a large margin. 
The model and training code can be found at our \href{https://github.com/phermosilla/msm}{Github} repository.
\end{abstract}

%% file: 1_intro.tex
\section{Introduction}
\label{sec:intro}

2D self-supervised models, such as DINOv2~\cite{oquab2023dinov2}, have become an integral part of modern computer vision. 
These models are typically pre-trained on large unlabeled datasets, providing task-agnostic features that can be used off-the-shelf to solve any computer vision task without the need for fine-tuning, making them particularly useful in scenarios with limited data~\cite{gui2024survey}.
In contrast to 2D computer vision, the field of 3D scene understanding lacks comparable models.
Instead, it relies on consolidating features from 2D foundation models into 3D~\cite{Peng2023OpenScene,takmaz2023openmask3d,man2024lexicon3d} or using them in a 2D-3D knowledge distillation setup~\cite{zhu2023ponderv2, chen2023bridging, zhang2024condense}.

Although this is still an emerging field, self-supervised methods for processing 3D scenes have started to gain traction, with several approaches emerging in recent years~\cite{PointContrast2020, hou_csc_2021, wu_masked_2023, Xu_2023_CVPR, wu_mitigating_2024, wang_groupcontrast_2024}.
Some of these methods incorporate \ac{MIM} objectives~\cite{MaskedAutoencoders2021,Xie_2022_CVPR,zhou2021ibot,oquab2023dinov2} in their 3D scene-based frameworks~\cite{wu_masked_2023,Xu_2023_CVPR}, where the model is tasked with reconstructing the input scene from a partial view.
In the 2D domain, such learning objectives have been shown to lead to semantically rich features better suited for dense prediction tasks~\cite{oquab2023dinov2}, achieving unprecedented performance for off-the-shelf feature evaluations.
Unfortunately, self-supervised learning on 3D scenes has so far failed to exhibit such semantic properties off-the-shelf, and it is used as a weight initialization step before fine-tuning the model in a downstream task.
We believe this is due to two main limitations: \textbf{(i)} The lack of a systematic protocol specific to 3D scene understanding to evaluate the representations learned by such models, and \textbf{(ii)} the lack of an effective 3D-scene specific masked prediction objective that takes into account the hierarchical nature of these models.
In this paper, we aim to address these limitations:

\textbf{(i)} First, we advocate that, to advance the field, we necessitate an evaluation protocol to directly evaluate the quality of the representations learned by self-supervised methods tailored explicitly to 3D scenes.
Models designed for 3D scene understanding have a hierarchical nature, usually following a UNet~\cite{ronneberger2015u} design.
Naively using the output of the last layer of such hierarchical models for off-the-shelf feature evaluation might not reflect the underlying semantic capabilities of the self-supervised model.
In a supervised setup, such models discard unnecessary information for the downstream task during decoding, whereas in a self-supervised setup, this information may be relevant for producing task-agnostic features.
To address this, we use tri-linear interpolation to upsample the feature maps of each decoder level and combine them to create a final task-agnostic feature map.
The resulting set of features retains the hierarchical information otherwise lost during decoding and can then be used for off-the-shelf feature evaluation in a linear probing or nearest-neighbor setup.
This evaluation reflects the effectiveness of representations learned by self-supervised models better than a fine-tuning protocol does, since the quality of the features is not masked by further optimization on the downstream task.
In a pilot study, we demonstrate that our hierarchical evaluation approach more effectively reveals the off-the-shelf feature capabilities of self-supervised methods. 
Furthermore, this study reveals a significant performance gap between supervised and self-supervised training, highlighting the necessity for a framework better suited for 3D scenes.

\textbf{(ii)} To address this gap, we propose a hierarchical self-supervised framework based on the \ac{MIM} objective specifically designed for models that process 3D scenes.
We argue that the failure of existing \ac{MIM} approaches to learn semantically relevant features from self-supervised learning alone is rooted in their design choices:
Some methods mask the input features of the 3D points~\cite{wu_masked_2023}, letting the model infer those from the geometry, simplifying the masking objective.
Additionally, some methods use the reconstruction of input features~\cite{wu_masked_2023,Xu_2023_CVPR} as training objective instead of deep features, which leads to reduced semantic information~\cite{assran2023ijepa}.
Lastly, existing methods do not consider the hierarchical nature of their models when designing the reconstruction loss~\cite{wu_masked_2023}.
With our proposed approach, \emph{Masked Scene Modeling}, we aim to overcome these limitations by making several crucial design choices to better learn the semantic properties of the 3D scene:
We perform a bottom-up hierarchical masking approach, where the encoder receives a masked sparse voxelization of the scene.
During decoding, the masked patches are included, and the model reconstructs the deep features of these patches obtained from a teacher model.
This design allows for a hierarchical reconstruction of the scene without information leakage from the geometric cues present in sparse representations.
Moreover, inferring deep features leads to a fast learning process with richer semantic features thanks to the use of abstract prediction targets~\cite{assran2023ijepa}.
Our extensive evaluation demonstrates that our self-supervised features can be used off-the-shelf to solve several tasks, achieving competitive performance, for the first time, when compared to supervised methods and significantly outperforming other existing self-supervised methods for 3D scenes.

%% file: 2_related_work.tex
\section{Related Work}
\label{sec:relatedwork}

\paragraph{Self-supervised methods for 3D scene understanding.}
In 3D self-supervised learning, there are two main lines of work: methods focused on pretext tasks designed for shapes representing single objects~\cite{hassani2019unsupervised,sauder2019sslshapes,Yu_2022_CVPR,pang2022PointMAE,zhang2022point,yan2024featpred}, and methods with self-supervised objectives designed for large 3D scenes composed of multiple objects~\cite{xie_pointcontrast_2020,hou_csc_2021,wu_masked_2023,chen_4D_Contrast,wang_groupcontrast_2024}.
While single object pretext objectives are able to perform well on object-centric tasks such as classification or segmentation~\cite{Yu_2022_CVPR,pang2022PointMAE,zhang2022point,yan2024featpred}, as shown experimentally by Xie et al.~\cite{xie_pointcontrast_2020}, they fall behind on complex 3D scene understanding tasks.
Xie et al.~\cite{xie_pointcontrast_2020} proposed one of the first scene-centric self-supervised methods, which employed a contrastive learning objective~\cite{chen2020simclr} at the point level.
This work was later improved by Hou et al.~\cite{hou_csc_2021} by partitioning the space around the points and using those to select meaningful negative examples for contrastive learning loss.
Chen et al.~\cite{chen_4D_Contrast} further extended the same idea to work with object trajectories within a scene.
Contrastive learning was also used by Zhang et al.~\cite{zhang_self-supervised_2021} and Huang et al.~\cite{huang_spatio-temporal_2021} at scene level using a momentum encoder as target.
Recently, Wang et al.~\cite{wang_groupcontrast_2024} have also suggested using an over-segmentation step to group points and a prototype clustering step to improve contrastive learning. 
However, in recent works, following the trends in 2D vision, the paradigm has shifted, and several methods have suggested using \ac{MIM} as pretext task~\cite{wu_masked_2023,Xu_2023_CVPR}.
Wu et al.~\cite{wu_masked_2023} combines per-point contrastive objective with color and normal reconstruction, while Xu et al.~\cite{Xu_2023_CVPR} aims at reconstructing neighboring point coordinates at different scales.
Despite these advancements, the reconstruction objectives in these works are limited to local features such as color or normals or to reconstructing neighboring point coordinates from sparse representations, which leads to features with lower semantic capabilities~\cite{assran2023ijepa}.
In contrast, this work advocates for deep abstract feature reconstruction of large masked areas in a hierarchical bottom-up manner, making the self-supervised model obtain semantically richer features at different scales.

\paragraph{Validation of self-supervised models.}
Early work used pre-trained models as weight initialization for downstream tasks and measured the increase in performance~\cite{vincent2008denoisingae}.
Current state-of-the-art models for images, on the other hand, evaluate the feature space directly by freezing the pre-trained model and using \ac{NN}~\cite{Caron_2021_dino,oquab2023dinov2,zhou2021ibot} and linear probing protocols to solve classifications tasks~\cite{chen2020simclr,Caron_2021_dino,oquab2023dinov2,zhou2021ibot}.
Although these protocols can evaluate the representation learning capabilities of the models better than simple fine-tuning, these are not commonly used in scene-centric 3D self-supervised learning.
Methods that use self-supervised learning from 3D scenes are usually evaluated by a fine-tuning protocol on different 3D scene understanding tasks~\cite{xie_pointcontrast_2020,hou_csc_2021,wu_masked_2023,Xu_2023_CVPR}.
Recent work~\cite{Xu_2023_CVPR} used a linear probing setup in their evaluation; however, this approach was not been the primary metric used to measure performance and they only relied on features of the last layer of hierarchical models. 
In this work, we propose an evaluation protocol that uses hierarchical features and better reflects the quality of the learned representations.

%% file: 3_metrics.tex
\section{Feature Evaluation Protocol}
\label{sec:metrics}

\begin{figure}
    \centering
    \includegraphics[width=\linewidth]{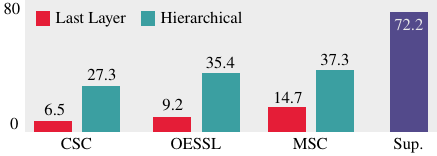}
    \caption{\textbf{Pilot study.} Our hierarchical features uncover better performance in all self-supervised models.
    Moreover, our study shows that existing approaches exhibit a large performance gap between supervised and self-supervised training.}
    \label{fig:pilot}
\end{figure}

In this section, first, we describe the proposed evaluation protocol designed to measure the quality of the representations learned by self-supervised hierarchical models, followed by a pilot study on existing self-supervised methods.

\paragraph{Hierarchical feature extraction.}
In a UNet~\cite{ronneberger2015u}-like architecture, like the one commonly used in 3D scene understanding, a hierarchical decoder reduces the feature dimensionality at each level while increasing the spatial resolution.
In supervised learning, the final layer is composed of a small number of features that contain the relevant information to solve the downstream task since the model can learn them from deeper levels and discard unnecessary information along the process.
In self-supervised learning, on the other hand, where the model should generate general features to solve various tasks, evaluating only the features of the last layer in the decoder might limit the information available and might discard valuable information within deeper levels.
\begin{wrapfigure}{r}{0.25\textwidth}
    \vspace{0cm}
    \centering
    \includegraphics[width=\linewidth]{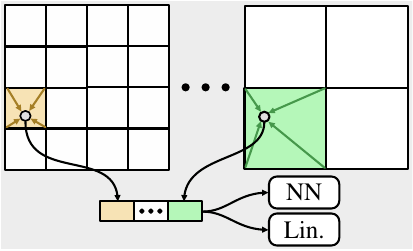}
    \caption{\textbf{Hierarchical features}}
    \label{fig:eval_protocol}
    \vspace{-.25cm}
\end{wrapfigure}
Therefore, in this paper, we suggest using a concatenation of the output features of each level in a hierarchical decoder, thus obtaining features with information at different scales.
In particular, we propose to use tri-linear interpolation, as shown in Figure~\ref{fig:eval_protocol}, to obtain distinct features that better reflect the semantic capabilities for each point in space.
These features can then be used off-the-shelf to solve downstream tasks.

\paragraph{Pilot study.}
We conducted a pilot study to validate the assumption that our hierarchical features are better suited for evaluating self-supervised models.
We collected pre-trained models from recent self-supervised methods~\cite{hou_csc_2021,wu_masked_2023,wu_mitigating_2024} that employed the same sparse convolution architecture and evaluated their linear probing performance on the downstream task of semantic segmentation on the ScanNet dataset~\cite{dai2017scannet}.
In particular, we compare two feature extraction approaches: Naively using only features from the last layer of the decoder or extracting hierarchical features using tri-linear interpolation.
Figure~\ref{fig:pilot} shows that only using the features of the last layer does not fully capture the semantic capabilities of the models, resulting in poor segmentation performance.
However, when the hierarchical features of all layers are used for linear probing, the model's ability to produce semantic features is much better captured, since the performance significantly increases.

From this study, we arrived at two main conclusions.
First, we confirmed our assumption that deeper layers of self-supervised models still contain relevant information lost during hierarchical decoding and can assist in solving a downstream task.
Second, the gap between supervised and self-supervised models is still large, limiting the application of existing self-supervised strategies in practice.
This highlights the necessity of new self-supervised approaches for 3D scene understanding that consider the hierarchical nature of the models used.
Based on these observations, in the following section, we describe our novel framework that uses self-supervision at different levels to better capture semantic relations in its features and can achieve supervised-level performance when hierarchical features are used off-the-shelf to solve various downstream tasks.

%% file: 4_methods.tex
\begin{figure*}
    \centering
    \includegraphics[width=\linewidth]{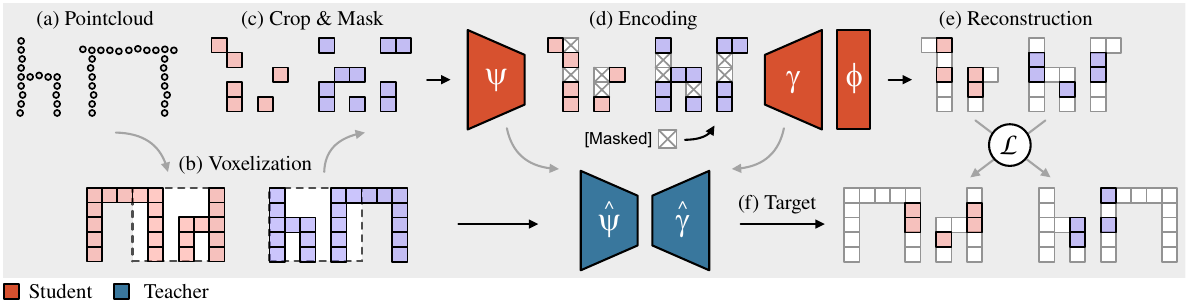}
    \caption{\textbf{Overview.} Our method receives as input a 3D scene represented as a pointcloud, \textbf{(a)}. The scene is voxelized into two different views, \textbf{(b)}, and then further cropped and masked, \textbf{(c)}. The student model first encodes the cropped views and then adds the masked voxels with a learnable token, \textbf{(d)}. The decoder processes the cropped views and reconstructs deep features of the masked tokens, \textbf{(e)}. The loss is computed in a cross-view manner where the target features, \textbf{(f)}, are obtained from a teacher model updated with EMA.}
    \label{fig:overview}
\end{figure*}

\section{Masked Scene Modeling}
\label{sec:methods}

This section introduces our self-supervised framework, named \emph{Masked Scene Modeling}, designed based on the findings of our pilot study in Section~\ref{sec:metrics} and tailored explicitly for 3D scene understanding.
First, Section~\ref{subsec:overview} presents the main components of our self-supervised framework.
Then, in Section~\ref{subsec:hierarchical} we describe in detail the hierarchical reconstruction objective at the core of our method.
Figure~\ref{fig:overview} presents an illustration of the proposed framework.

\subsection{Self-Supervised Training}
\label{subsec:overview}

The main self-supervised objective of our framework is: From a masked partial view of a scene, the model is tasked to reconstruct deep features given by a teacher model that has access to a view of the whole scene.
This objective not only forces the model to learn view-invariant features but also makes the model acquire a deep understanding of the scene's composition.
Our framework has five main components: view generation, feature encoding, feature decoding, reconstruction objective, and teacher model.

\paragraph{View generation.}
Our framework receives as input a 3D scene represented as a pointcloud, $\mathcal{P}$.
First, two different data augmentations are applied to $\mathcal{P}$ and the resulting scenes are then voxelized into $\mathcal{V}_1$ and $\mathcal{V}_2$, where only occupied voxels are stored in memory.
Then, $\mathcal{V}_1$ and $\mathcal{V}_2$ are further cropped to obtain a partial view of each scene, $\mathcal{C}_1$ and $\mathcal{C}_2$.
From each crop, we then randomly mask certain areas resulting in two sets of voxels, unmasked, $C_{v1}$ and $C_{v2}$, and masked voxels, $C_{m1}$ and $C_{m2}$.
These cropped views serve as input to our student model whilst the full voxelized views are given to the teacher model.

\paragraph{Feature encoding.}
The proposed framework assumes a hierarchical model composed of an encoder $\Psi$ and a decoder $\gamma$.
Our framework, first, encodes the unmasked voxels $C_{v1}$ and $C_{v2}$ using $\Psi$, resulting in a set of voxel features $F^e_{v1} = \Psi(C_{v1})$ and $F^e_{v2} = \Psi(C_{v2})$.

\paragraph{Feature decoding.}
Before decoding the features $F^e_{v1}$ and $F^e_{v2}$ of unmasked voxels with $\gamma$, we incorporate the features from masked voxels, $F^e_{m1}$ and $F^e_{m2}$, by assigning them a learnable token, $T$.
This process results in the feature maps $F^e_{1} = F^e_{v1} \cup F^e_{m1}$ and $F^e_{2} = F^e_{v2} \cup F^e_{m2}$.
The combined features are then processed with the decoder $\gamma$, which generates the decoded features for each partial view, $F^d_{1} = \gamma(F^e_{1})$ and $F^d_{2} = \gamma(F^e_{2})$.
%Once trained, these features will be the output of our model.

\paragraph{Reconstruction objective.}
The self-supervised objective in our framework is the reconstruction of deep features of the masked voxels.
Therefore, from the decoded features $F^d_{1}$ and $F^d_{2}$, we select those belonging to masked voxels, $F^d_{m1}$ and $F^d_{m2}$, and process them with a predictor model, $\Phi$, implemented as a small \ac{MLP}.
This results in the predicted features $F^p_{m1} = \Phi(F^d_{m1})$ and $F^p_{m2} = \Phi(F^d_{m2})$.
The target features used for supervision are obtained by processing the full scene views with a teacher encoder and decoder, $\hat{F}^d_{1} = \hat{\gamma}(\hat{\Psi}(\mathcal{V}_1))$ and $\hat{F}^d_{2} = \hat{\gamma}(\hat{\Psi}(\mathcal{V}_2))$.
This objective enforces the model to infer semantic knowledge of the full scene from only a cropped and masked scene.
In addition, to obtain view-invariant features, we perform cross-reconstruction between views, resulting in the following reconstruction loss:

\begin{equation}
    \mathcal{L} = | F^p_{m1} - \hat{F}^d_{m2} | + | F^p_{m2} - \hat{F}^d_{m1} |
\end{equation}

\paragraph{Teacher model.}
Following common practices of self-supervised methods for images~\cite{he2019moco,zhou2021ibot,oquab2023dinov2}, we use as our teacher a model with the same architecture as the student, but whose parameters are updated as the \ac{EMA} of the parameters of the student.
The slow update of the teacher parameters reduces feature variation during training, making the self-distillation process more robust~\cite{he2019moco}.
This makes the model learn rich semantic features and avoids a common problem of self-supervised methods, mode collapse, where the model learns to predict always the same feature vector independently of the input.

\subsection{Hierarchical Reconstruction}
\label{subsec:hierarchical}

\begin{figure}
    \centering
    \includegraphics[width=\linewidth]{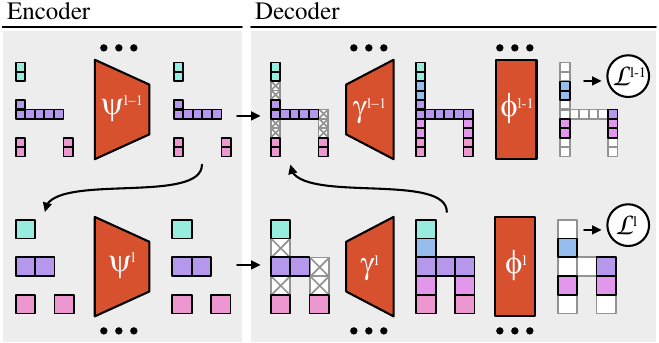}
    \caption{\textbf{Hierarchical reconstruction.} The masked voxelization is processed by our hierarchical encoder. The decoder processes the encoded features in a bottom-up manner by first including the masked voxels with a learnable token. Each level is used in the loss computation before the decoded features are upscaled and combined with the skip connection from the previous level.}
    \label{fig:hier-rec}
\end{figure}

One of the key insights from our pilot study in Section~\ref{sec:metrics} is that all levels in the hierarchical model carry relevant information that can be used in a downstream task.
Therefore, in this paper, we suggest performing the reconstruction at each level in the hierarchical model to learn features at different scales.
Figure~\ref{fig:hier-rec} illustrates this process in detail.

When our encoder processes a scene, we receive a feature map of the unmasked voxels for each level, $F^e_v = (F^{e1}_v, ..., F^{eL}_v)$, where $L$ is the number of levels in the hierarchical encoder.
These features are then combined with the features of masked voxels in each level by assigning a different learnable token in each level, $T=(T^1, ..., T^L)$, resulting in the final encoder features, $F^e = (F^{e1}, ..., F^{eL})$.
The decoder then starts processing the combined features of the last layer $L$ and generates the decoded features for the last layer, $F^{dL}$.
From these features, we compute the predicted features for the masked voxels at this level using the predictor network, $\Phi^L$.
Features $F^{dL}$ are then upsampled and combined with the features of the encoder at the previous level, $F^{eL-1}$,  using a skip connection.
The process is repeated using different predictor networks at each level, $\Phi=(\Phi^1, ..., \Phi^L)$, until we reach the initial voxel resolution.
Then, the predicted features at each level are supervised with the features of our teacher model:

\begin{equation}
    \mathcal{L} = \sum_{l=0}^{L}| F^{pl}_{m1} - \hat{F}^{dl}_{m2} | + | F^{pl}_{m2} - \hat{F}^{dl}_{m1} |
\end{equation}

\paragraph{Bottom-up vs Top-down reconstruction.}
The presented hierarchical reconstruction performs reconstruction in a bottom-up manner, including the masked token only on the decoder while completely removing masked tokens from the encoder.
Another possible approach could be to include such learnable tokens in the encoder, similar to Wu \etal~\cite{wu_masked_2023}.
Unfortunately, this will allow deeper levels to infer geometric information of the masked regions from previous levels using features from neighboring voxels, making the self-supervised task easier and, therefore, as we show in our ablation studies, leading to lower performance.

\paragraph{Masking.}
One of the key components in our framework is the masking of random voxels.
In order to avoid information leakage between levels, we perform consistent masking between levels in the hierarchy, \ie the same areas are masked at different voxel resolutions.
Therefore, to mask voxels in the lowest levels in the hierarchy, where each voxel covers a large area of the scene, we fix the mask patch size to the voxel resolution of this level, resulting in large independent masked areas.
Recent work on \ac{MIM}~\cite{xie2021simmim} has shown that this strategy leads to a more stable performance for different masking ratios.

%% file: 5_comp_sota.tex
\section{Main Results}
\label{sec:mainresults}

In this section, we present extensive experiments where we evaluate the representations learned by our self-supervised model on common 3D scene understanding tasks.
For a detailed description, additional experiments, and ablation studies, we refer the reader to the supplementary material.

\subsection{Experimental Setup}

\paragraph{Baselines.}
We compare our self-supervised model to recent self-supervised models for 3D scenes trained exclusively with 3D data for which code and pre-trained weights were available at the time of the submission.
These cover training objectives based on contrastive learning, masked point modeling, and clustering-based approaches:

\begin{itemize}
    \item \textbf{CSC}~\cite{hou_csc_2021}. This CVPR 2021 work proposed a contrastive learning approach with a carefully designed sampling strategy of positive and negative points.

    \item \textbf{MSC}~\cite{wu_masked_2023}. This work was presented at CVPR 2023 and proposed a contrastive learning approach combined with masking and reconstructing point colors and normals.

    \item \textbf{MM3D}~\cite{Xu_2023_CVPR}. This work was also presented at CVPR 2023 and it suggested a masking strategy and a hierarchical reconstruction of point coordinates combined with self-distillation in non-masked regions.

    \item \textbf{OESSL}~\cite{wu_mitigating_2024}. This CVPR 2024 work used a clustering approach to perform data augmentation of plausible objects in the scene combined with contrastive learning.
\end{itemize}

For a fair evaluation, we downloaded the weights of all the models trained by the authors and used the resulting features without any fine-tuning.
Additionally, we trained our model with the best-performing baseline, MSC.
We also provide two models trained from scratch in a supervised fashion on the downstream tasks for comparison.

\paragraph{Pre-Training.}
We follow our baselines~\cite{hou_csc_2021,wu_masked_2023,Xu_2023_CVPR,wu_mitigating_2024} and pre-train our model only on the ScanNet dataset~\cite{dai2017scannet} with a masking ratio of $0.4$.
We train our model for $1800$ epochs on a computer with 4 $\times$ A6000 GPUs for 3 days.

\paragraph{Model.}
Our model uses a UNet architecture~\cite{ronneberger2015u} with 2 ResNet blocks in each level of the encoder and decoder using sparse convolutions~\cite{3DSemanticSegmentationWithSubmanifoldSparseConvNet}.
In the last two levels of encoder and decoder, similar to Stable Diffusion models~\cite{rombach2021highresolution}, we also incorporate two \ac{MHA} blocks with a serialization strategy as in PTv3~\cite{wu2024ptv3}.
However, we remove the xCPE layers and use window attention instead of block attention.
We refer to this architecture as Hybrid UNet (HUNet).

%%%%%%%%%%%%%%%%%%%%%%%%% SEM SEG

\subsection{Semantic Segmentation}

Semantic segmentation in 3D scenes aims to predict the class of each point in the scene from a closed set of classes.
Successfully solving such a task indicates that the features used contain semantically rich information.

\paragraph{Datasets.}
We evaluate all models on three different datasets, ScanNet~\cite{dai2017scannet}, ScanNet200~\cite{rozenberszki2022language}, and S3DIS~\cite{armeni2017joint}.
All datasets are composed of dense 3D scans of indoor scenes with objects from $20$, $200$, and $13$ different classes, respectively.
We follow the standard splits for ScanNet and ScanNet200, reporting \ac{mIoU} performance on the validation set and reporting performance on the Area5 for the S3DIS dataset.
For the evaluation, we use two protocols.
In the first one, \ac{NN}, the class of each point in the validation set is predicted by searching the point in the training set with the most similar feature and using its class as the predictor.
Since the scenes are composed of a large number of points, to reduce the time of the \ac{NN} search, we group points in super-points similar to Rozenberszki \etal~\cite{rozenberszki2024unscene3d}, and perform the similarity search at the super-point level.
The second evaluation protocol, Linear, trains a linear layer on top of the off-the-shelf features.

\begin{table}
\centering
\setlength{\tabcolsep}{7.5pt}
\small
\begin{tabular}{llcccc}
    \toprule
    & & BB & ScN & ScN200 & S3DIS \\

    \midrule

    \cbg & \cbg  & \footnotesize \cbg \textcolor{SupColor}{SR-UNet} & \cbg \textcolor{SupColor}{72.2} & \cbg \textcolor{SupColor}{25.0} & \cbg \textcolor{SupColor}{68.2} \\
    \cbg  &   \multirow{-2}{*}{\cbg \textcolor{SupColor}{Supervised}} & \footnotesize \cbg \textcolor{SupColor}{HUNet} & \cbg \textcolor{SupColor}{77.0} & \cbg \textcolor{SupColor}{35.4} & \cbg \textcolor{SupColor}{71.3} \\

    \midrule
    
    \multirow{6}{*}{\rotatebox[origin=c]{90}{NN}} & MM3D~\cite{Xu_2023_CVPR} & \footnotesize PT & 19.3 & 3.9 & 24.6 \\
    & CSC~\cite{hou_csc_2021} & \footnotesize SR-UNet & 24.1 & 4.6 & 32.0 \\
    & OESSL~\cite{wu_mitigating_2024} & \footnotesize SR-UNet & 31.0 & 6.6 & 34.0 \\
    & \multirow{2}{*}{MSC~\cite{wu_masked_2023}} & \footnotesize SR-UNet & 31.3 & 7.1 & 35.2 \\
    & & \footnotesize HUNet & 39.9 & 11.0 & 39.7 \\

    \cmidrule{2-6}

    & Ours & \footnotesize HUNet & \textbf{65.7} & \textbf{22.7} & \textbf{45.7} \\
    
    \midrule

    \multirow{6}{*}{\rotatebox[origin=c]{90}{Linear}} & MM3D~\cite{Xu_2023_CVPR} & \footnotesize PT & 26.7 & 4.6 & 36.6 \\
    & CSC~\cite{hou_csc_2021} & \footnotesize SR-UNet & 27.3 & 5.6 & 31.1 \\
    & OESSL~\cite{wu_mitigating_2024} & \footnotesize SR-UNet & 35.4 & 9.1 & 35.8 \\
    & \multirow{2}{*}{MSC~\cite{wu_masked_2023}} & \footnotesize SR-UNet & 37.3 & 10.2 & 41.7 \\
    & & \footnotesize HUNet & 58.2 & 20.3 & 57.7 \\

    \cmidrule{2-6}

    & Ours & \footnotesize HUNet & \textbf{68.7} & \textbf{26.8} & \textbf{59.5} \\
    
    \bottomrule

\end{tabular}
\caption{\textbf{Semantic Segmentation.} Performance of different self-supervised models on the task of semantic segmentation (mIoU).}
\label{tbl:semseg-results}
\end{table}

\paragraph{Results.}
Tbl.~\ref{tbl:semseg-results} shows the result of our experiments.
From the \ac{NN} protocol, we can see that our features are able to achieve much better performance than existing models, outperforming them by more than \textbf{+30} points in ScanNet, \textbf{+15} points on ScanNet200, and \textbf{+10} points on S3DIS.
When we compare our model with the same architecture trained with MSC, we can see that it surpasses it by a large margin, obtaining improvements of \textbf{+25}, \textbf{+11}, and \textbf{+6}.
For the Linear evaluation protocol, we can see similar improvements.
We outperform existing models by \textbf{+30}, \textbf{+16}, and \textbf{+18} points.
When we compare our architecture trained with MSC, we can still see large improvements in the performance of \textbf{+10}, \textbf{+6}, and \textbf{+2}.
Lastly, Tbl.~\ref{tbl:semseg-results} also shows that, when compared to supervised methods trained from scratch, existing self-supervised models achieve significantly lower performance on all datasets.
On the other hand, our model can achieve competitive performance, and even surpass, models trained from scratch, further underlying the semantic relations captured in our off-the-shelf features.

%%%%%%%%%%%%%%%%%%%%%%%%% INST SEG

\subsection{Instance Segmentation}

The task of instance segmentation is more challenging since it requires the prediction of the semantic class of each point in the scene and the mask of each independent object instance.
Successfully solving such a task will indicate that the model is not only aware of the semantics of the scene but also contains object-aware features.

\paragraph{Datasets.}
As our datasets, we use again ScanNet~\cite{dai2017scannet}, ScanNet200~\cite{rozenberszki2022language}, and S3DIS~\cite{armeni2017joint}.
We evaluate the performance of the models on those datasets with \ac{mAP} with a threshold of $0.5$.
Our evaluation protocol uses a linear layer on top of the frozen features to predict the semantic class and a single layer \ac{MLP} to predict the displacement vector in the PointGroup algorithm~\cite{jiang2020pointgroup}.

\begin{table}
\centering
\setlength{\tabcolsep}{7.5pt}
\small
\begin{tabular}{llcccc}
    \toprule
    & & BB & ScN & ScN200 & S3DIS \\

    \midrule
    
    \cbg & \cbg  & \footnotesize \cbg \textcolor{SupColor}{SR-UNet} & \cbg \textcolor{SupColor}{56.9} & \cbg \textcolor{SupColor}{24.5} & \cbg \textcolor{SupColor}{59.3} \\
    \cbg  &   \multirow{-2}{*}{\cbg \textcolor{SupColor}{Supervised}} & \footnotesize \cbg \textcolor{SupColor}{HUNet} & \cbg \textcolor{SupColor}{65.5} & \cbg \textcolor{SupColor}{32.8} & \cbg \textcolor{SupColor}{58.4} \\
    
    \midrule

    \multirow{6}{*}{\rotatebox[origin=c]{90}{Linear}} & MM3D~\cite{Xu_2023_CVPR} & \footnotesize PT & 4.3 & 0.4 & 7.7 \\
    & CSC~\cite{hou_csc_2021} & \footnotesize SR-UNet & 3.5 & 0.1 & 14.8 \\
    & OESSL~\cite{wu_mitigating_2024} & \footnotesize SR-UNet & 13.6 & 2.5 & 16.5 \\
    & \multirow{2}{*}{MSC~\cite{wu_masked_2023}} & \footnotesize SR-UNet & 10.1 & 1.6 & 16.6 \\
    & & \footnotesize HUNet & 24.5 & 3.8 & 18.2 \\

    \cmidrule{2-6}

    & Ours & \footnotesize HUNet & \textbf{44.4} & \textbf{8.8} & \textbf{23.2} \\
    
    \bottomrule

\end{tabular}
\caption{\textbf{Instance Segmentation.} Performance of self-supervised models on the task of instance segmentation (mAP@50).}
\label{tbl:instseg-results}
\end{table}

\paragraph{Results.}
Tbl.~\ref{tbl:instseg-results} presents the results of this experiment.
We can see that most of the competing methods struggle to solve this challenging task.
Our model, on the other hand, is able to outperform all models by more than \textbf{+30}, \textbf{+7}, and \textbf{+6} points on ScanNet, ScanNet200, and S3DIS respectively.
When compared to the HUNet trained with MSC, our method still maintains similar gains, and at the same time reduces the gap between supervised and self-supervised methods. 
With this, we show that our learned features not only represent semantic information effectively but also capture object-level properties.% better than all previous self-supervised methods.

%%%%%%%%%%%%%%%%%%%%%%%%% 3D VISUAL GROUNDING

\subsection{3D Visual Grounding}
The task of 3D visual grounding places high importance on object-level reasoning, where the model has to locate an object in the scene from a text description.
This task can be divided into two subtasks: object detection and object discrimination, where the model selects the appropriate instance based on the text description.
Since object detection capabilities were evaluated in the previous experiments, we follow~\cite{man2024lexicon3d} and only evaluate the object discriminator task by using ground truth boxes of all objects in the scene.

\paragraph{Datasets.}
We use the ScanRefer~\cite{chen2020scanrefer} dataset, which provides text descriptions of objects from different 3D scenes.
We report accuracy on the three evaluation sets with different difficulty levels, \emph{Unique}, \emph{Multiple}, and \emph{Overall}.
As our evaluation protocol, following~\cite{man2024lexicon3d}, we use a small model composed of self- and cross-attention layers between the object features (obtained by averaging voxel features inside the bounding boxes) and the text embeddings.

\begin{table}
\centering
\setlength{\tabcolsep}{9pt}
\small
\begin{tabular}{llcccc}
    \toprule
    & & BB & U & M & O \\

    \midrule
    
    \cbg & \cbg  & \footnotesize \cbg \textcolor{SupColor}{SR-UNet} & \cbg \textcolor{SupColor}{78.8} & \cbg \textcolor{SupColor}{35.9} & \cbg \textcolor{SupColor}{44.3} \\
    \cbg  &   \multirow{-2}{*}{\cbg \textcolor{SupColor}{Supervised}} & \footnotesize \cbg \textcolor{SupColor}{HUNet} & \cbg \textcolor{SupColor}{--} & \cbg \textcolor{SupColor}{--} & \cbg \textcolor{SupColor}{--} \\
    
    \midrule

    \multirow{6}{*}{\rotatebox[origin=c]{90}{Cross-Att.}} & MM3D~\cite{Xu_2023_CVPR} & \footnotesize PT & 65.8 & 32.1 & 40.4 \\
    & CSC~\cite{hou_csc_2021} & \footnotesize SR-UNet & 66.9 & 32.6 & 39.1 \\
    & OESSL~\cite{wu_mitigating_2024} & \footnotesize SR-UNet & 73.8 & 35.6 & 43.0 \\
    & \multirow{2}{*}{MSC~\cite{wu_masked_2023}} & \footnotesize SR-UNet & 77.1 & 37.4 & 45.0 \\
    & & \footnotesize HUNet & 84.4 & 42.9 & 51.0 \\

    \cmidrule{2-6}

    & Ours & \footnotesize HUNet & \textbf{87.1} & \textbf{44.3} & \textbf{55.7} \\
    
    \bottomrule

\end{tabular}
\caption{\textbf{3D Visual Grounding.} Accuracy of different self-supervised models on the task of 3D visual grounding.}
\label{tbl:3dvg-results}
\end{table}

\paragraph{Results.}
We present our results in Tbl.~\ref{tbl:3dvg-results}, which shows that existing self-supervised models are able to achieve certain moderate accuracy and, in some cases, even surpass the supervised model.
However, our model exhibits significantly improved performance, outperforming them all by large margins of \textbf{+10}, \textbf{+7}, and \textbf{+10} on the validation sets \emph{Unique}, \emph{Multiple}, and \emph{Overall}.
Our Hybrid model trained with MSC is also able to provide an improvement over existing models but still falls behind our proposed self-supervised objective.
Unfortunately, training HUNet from scratch leads to unstable training, being unable to converge.

%%%%%%%%%%%%%%%%%%%%%%%%% LIMITED ANNOTATION

\begin{table}
\centering
\setlength{\tabcolsep}{4.5pt}
\small
\begin{tabular}{llcccccc}
    \toprule
    & & BB & 1\% & 5\% & 10\% & 20\% & 100\%\\

    \midrule
    
    \cbg & \cbg  & \footnotesize \cbg \textcolor{SupColor}{SR-UNet} & \cbg \textcolor{SupColor}{26.1} & \cbg \textcolor{SupColor}{47.8} & \cbg \textcolor{SupColor}{56.7} & \cbg \textcolor{SupColor}{62.9}  & \cbg \textcolor{SupColor}{72.2} \\
    \cbg  &   \multirow{-2}{*}{\cbg \textcolor{SupColor}{Supervised}} & \footnotesize \cbg \textcolor{SupColor}{HUNet} & \cbg \textcolor{SupColor}{19.7} & \cbg \textcolor{SupColor}{36.4} & \cbg \textcolor{SupColor}{52.5} & \cbg \textcolor{SupColor}{67.1} & \cbg \textcolor{SupColor}{77.0} \\
    
    \midrule
    
    \multirow{6}{*}{\rotatebox[origin=c]{90}{NN}} & MM3D~\cite{Xu_2023_CVPR} & \footnotesize PT & 9.2 & 13.0 & 14.1 & 15.7 & 19.3 \\
    & CSC~\cite{hou_csc_2021} & \footnotesize SR-UNet & 12.9 & 16.2 & 18.7 & 19.9 & 24.1 \\
    & OESSL~\cite{wu_mitigating_2024} & \footnotesize SR-UNet & 15.9 & 20.8 & 23.7 & 25.9 & 31.0 \\
    & \multirow{2}{*}{MSC~\cite{wu_masked_2023}} & \footnotesize SR-UNet & 17.6 & 21.6 & 25.0 & 26.8 & 31.3 \\
    & & \footnotesize HUNet & 22.0 & 30.6 & 33.0 & 35.1 & 39.9 \\

    \cmidrule{2-8}

    & Ours & \footnotesize HUNet & \textbf{35.2} & \textbf{53.6} & \textbf{59.4} & \textbf{60.8} & \textbf{65.7} \\
    
    \midrule

    \multirow{6}{*}{\rotatebox[origin=c]{90}{Linear}} & MM3D~\cite{Xu_2023_CVPR} & \footnotesize PT & 16.4 & 22.6 & 25.5 & 26.6 & 26.7 \\
    & CSC~\cite{hou_csc_2021} & \footnotesize SR-UNet & 17.1 & 21.2 & 26.0 & 27.3 & 27.3 \\
    & OESSL~\cite{wu_mitigating_2024} & \footnotesize SR-UNet & 20.4 & 28.5 & 32.7 & 34.4 & 35.4 \\
    & \multirow{2}{*}{MSC~\cite{wu_masked_2023}} & \footnotesize SR-UNet & 21.9 & 31.4 & 34.4 & 36.2 & 37.3 \\
    & & \footnotesize HUNet & 30.8 & 44.3 & 50.7 & 54.5 & 58.2 \\

    \cmidrule{2-8}

    & Ours & \footnotesize HUNet & \textbf{35.1} & \textbf{54.5} & \textbf{61.5} & \textbf{63.5} & \textbf{68.7} \\

    \bottomrule
\end{tabular}
\caption{\textbf{Efficiency benchmark.} Semantic segmentation performance with a limited number of scenes in the training set.}
\label{tbl:effi-results-1}
\end{table}

\begin{table}
\centering
\setlength{\tabcolsep}{5pt}
\small
\begin{tabular}{llcccccc}
    \toprule

    & & BB & 20 & 50 & 100 & 200 & Full\\

    \midrule
    
    \cbg & \cbg  & \footnotesize \cbg \textcolor{SupColor}{SR-UNet} & \cbg \textcolor{SupColor}{41.9} & \cbg \textcolor{SupColor}{53.9} & \cbg \textcolor{SupColor}{62.2} & \cbg \textcolor{SupColor}{65.5}  & \cbg \textcolor{SupColor}{72.2} \\
    \cbg  &   \multirow{-2}{*}{\cbg \textcolor{SupColor}{Supervised}} & \footnotesize \cbg \textcolor{SupColor}{HUNet} & \cbg \textcolor{SupColor}{62.7} & \cbg \textcolor{SupColor}{68.9} & \cbg \textcolor{SupColor}{73.1} & \cbg \textcolor{SupColor}{73.9} & \cbg \textcolor{SupColor}{77.0} \\
    
    \midrule
    
    \multirow{6}{*}{\rotatebox[origin=c]{90}{NN}} & MM3D~\cite{Xu_2023_CVPR} & \footnotesize PT & 10.9 & 11.8 & 12.9 & 13.5 & 19.3 \\
    & CSC~\cite{hou_csc_2021} & \footnotesize SR-UNet & 15.3 & 17.1 & 18.4 & 19.6 & 24.1 \\
    & OESSL~\cite{wu_mitigating_2024} & \footnotesize SR-UNet & 20.0 & 22.4 & 24.4 & 24.9 & 31.0 \\
    & \multirow{2}{*}{MSC~\cite{wu_masked_2023}} & \footnotesize SR-UNet & 20.3 & 22.4 & 23.6 & 25.1 & 31.3 \\
    & & \footnotesize HUNet & 25.4 & 27.6 & 30.0 & 32.0 & 39.9 \\

    \cmidrule{2-8}

    & Ours & \footnotesize HUNet & \textbf{55.9} & \textbf{60.3} & \textbf{61.3} & \textbf{62.7} & \textbf{65.7} \\
    
    \midrule

    \multirow{6}{*}{\rotatebox[origin=c]{90}{Linear}} & MM3D~\cite{Xu_2023_CVPR} & \footnotesize PT & 20.6 & 24.2 & 25.5 & 25.9 & 26.7 \\
    & CSC~\cite{hou_csc_2021} & \footnotesize SR-UNet & 23.6 & 25.6 & 26.5 & 26.5 & 27.3 \\
    & OESSL~\cite{wu_mitigating_2024} & \footnotesize SR-UNet & 30.1 & 32.2 & 33.2 & 33.7 & 35.4 \\
    & \multirow{2}{*}{MSC~\cite{wu_masked_2023}} & \footnotesize SR-UNet & 31.7 & 34.1 & 35.1 & 35.6 & 37.3\\
    & & \footnotesize HUNet & 51.9 & 55.3 & 56.3 & 56.8 & 58.2 \\

    \cmidrule{2-8}

    & Ours & \footnotesize HUNet & \textbf{62.9} & \textbf{65.9} & \textbf{67.3} & \textbf{68.1} & \textbf{68.7} \\
    
    \bottomrule

\end{tabular}
\caption{\textbf{Efficiency benchmark.} Semantic segmentation performance with a limited number of annotated points per scene.}
\label{tbl:effi-results-2}
\end{table}

\subsection{Limited annotations}

In this task, we evaluate the performance of the models under different numbers of annotations on the task of semantic segmentation.
This highlights the utility of self-supervised methods when data for the downstream task is scarce.

\paragraph{Datasets.}
We use the benchmarks proposed by Hou \etal~\cite{hou_csc_2021}, in which two protocols are used for evaluation.
In the first one, the number of annotated scenes for training is reduced to $1\,$\%, $5\,$\%, $10\,$\%, and $20\,$\% of the total scenes in ScanNet~\cite{dai2017scannet}.
In the second one, the number of annotated points per scene is reduced to $20$, $50$, $100$, and $200$.

\paragraph{Results.}
Tbl.~\ref{tbl:effi-results-1} presents the results when the number of scenes available for training is reduced.
The results show that our model is able to surpass, not only all other self-supervised methods but also all supervised methods when the number of training scenes is reduced to $10\,$\% of the original set using both evaluation protocols.
When the number of annotated points is reduced, Tbl.~\ref{tbl:effi-results-2} shows that our model is also able to outperform all other self-supervised methods.
When compared to supervised methods, our model is able to outperform the SR-UNet model in all benchmarks and achieves performance similar to that of the HUNet model.

%%%%%%%%%%%%%%%%%%%%%%%%% 2D FOUNDATION MODELS

\subsection{Comparison to 2D Foundation Models}

A common practice in 3D understanding, due to the lack of general 3D models, is to lift features from pre-trained 2D foundation models into 3D~\cite{Peng2023OpenScene,takmaz2023openmask3d,man2024lexicon3d}.
Therefore, in this experiment, we compare the performance of our self-supervised model to different 2D foundation models as done in Lexicon3D~\cite{man2024lexicon3d}.
We compare our model in the tasks of semantic segmentation and 3D visual grounding.

\begin{table}
\centering
\setlength{\tabcolsep}{4.5pt}
\small
\begin{tabular}{llccccc}
    \toprule
    & &  & \multicolumn{1}{c}{Sem. Seg.} & \multicolumn{3}{c}{3D VG} \\
    \cmidrule(l{2pt}r{2pt}){4-4}
    \cmidrule(l{2pt}r{2pt}){5-7}
    & & BB & ScN & U & M & O \\

    \midrule
    
    \cbg & \cbg  & \footnotesize \cbg \textcolor{SupColor}{SR-UNet} & \cbg \textcolor{SupColor}{72.2} & \cbg \textcolor{SupColor}{78.8} & \cbg \textcolor{SupColor}{35.9} & \cbg \textcolor{SupColor}{44.3} \\
    \cbg  & \multirow{-2}{*}{\cbg \textcolor{SupColor}{Supervised}} & \footnotesize \cbg \textcolor{SupColor}{HUNet} & \cbg \textcolor{SupColor}{77.0} & \cbg \textcolor{SupColor}{--} & \cbg \textcolor{SupColor}{--} & \cbg \textcolor{SupColor}{--} \\
    
    \midrule

    \multirow{4}{*}{\rotatebox[origin=c]{90}{2D}} & DINOv2~\cite{oquab2023dinov2} & \footnotesize ViT & 62.8 & 87.0 & 43.4 & 52.0 \\
    & LSeg~\cite{li2022languagedriven} & \footnotesize ViT & 47.5 & 88.1 & 41.2 & 50.4 \\
    & CLIP~\cite{radford2021learning} & \footnotesize ViT & 3.4 & 86.5 & 41.6 & 50.4 \\
    & SD~\cite{rombach2022high} & \footnotesize HUNet & 42.6 & 86.4 & 41.9 & 50.6 \\

    \midrule

    & Ours & \footnotesize HUNet & \textbf{68.7} & \textbf{87.1} & \textbf{44.3} & \textbf{52.7} \\
    
    \bottomrule

\end{tabular}
\caption{\textbf{2D Foundation models.} Comparison to 2D foundation models on semantic segmentation and 3D visual grounding.}
\label{tbl:2DF-results}
\end{table}

%\begin{table}
%\centering
%\setlength{\tabcolsep}{4.5pt}
%\small
%\begin{tabular}{llcccccc}
%    \toprule
%    & &  & \multicolumn{2}{c}{Sem. Seg.} & \multicolumn{3}{c}{3D VG} \\
%    \cmidrule(l{2pt}r{2pt}){4-5}
%    \cmidrule(l{2pt}r{2pt}){6-8}
%    & & BB & ScN & S3DIS & U & M & O \\
%    \midrule 
%    \cbg & \cbg  & \footnotesize \cbg \textcolor{SupColor}{SR-UNet} & \cbg \textcolor{SupColor}{72.2} & \cbg \textcolor{SupColor}{68.2} & \cbg \textcolor{SupColor}{78.8} & \cbg \textcolor{SupColor}{35.9} & \cbg \textcolor{SupColor}{44.3} \\
%    \cbg  & \multirow{-2}{*}{\cbg \textcolor{SupColor}{Supervised}} & \footnotesize \cbg \textcolor{SupColor}{HUNet} & \cbg \textcolor{SupColor}{77.0} & \cbg \textcolor{SupColor}{71.3} & \cbg \textcolor{SupColor}{--} & \cbg \textcolor{SupColor}{--} & \cbg \textcolor{SupColor}{--} \\   
%    \midrule
%    \multirow{4}{*}{\rotatebox[origin=c]{90}{2D}} & DINOv2~\cite{oquab2023dinov2} & \footnotesize ViT & 62.8 & -- & 87.0 & 43.4 & 52.0 \\
%    & LSeg~\cite{li2022languagedriven} & \footnotesize ViT & 47.5 & -- & 88.1 & 41.2 & 50.4 \\
%    & CLIP~\cite{park2020contrastive} & \footnotesize ViT & 3.4 & -- & 86.5 & 41.6 & 50.4 \\
%    & SD~\cite{rombach2022high} & \footnotesize HUNet & 42.6 & -- & 86.4 & 41.9 & 50.6 \\
%    \midrule
%    & Ours & \footnotesize HUNet & \textbf{68.7} & \textbf{59.5} & \textbf{87.1} & \textbf{44.3} & \textbf{52.7} \\   
%    \bottomrule
%\end{tabular}
%\caption{\textbf{2D Foundation models.} Comparison to 2D foundation models on semantic segmentation and 3D visual grounding.}
%\label{tbl:2DF-results}
%\end{table}

\paragraph{Results.}
Tbl.~\ref{tbl:2DF-results} presents the results of this experiment.
While 2D foundation models, can show impressive performance despite the domain gap, our 3D-native self-supervised model is able to outperform them all in all experiments, showing that representations learned natively in 3D better capture the 3D-specific properties of the scene.

%%%%%%%%%%%%%%%%%%%%%%%%% ABLATIONS

\begin{figure}
    \centering
    \includegraphics[width=\linewidth]{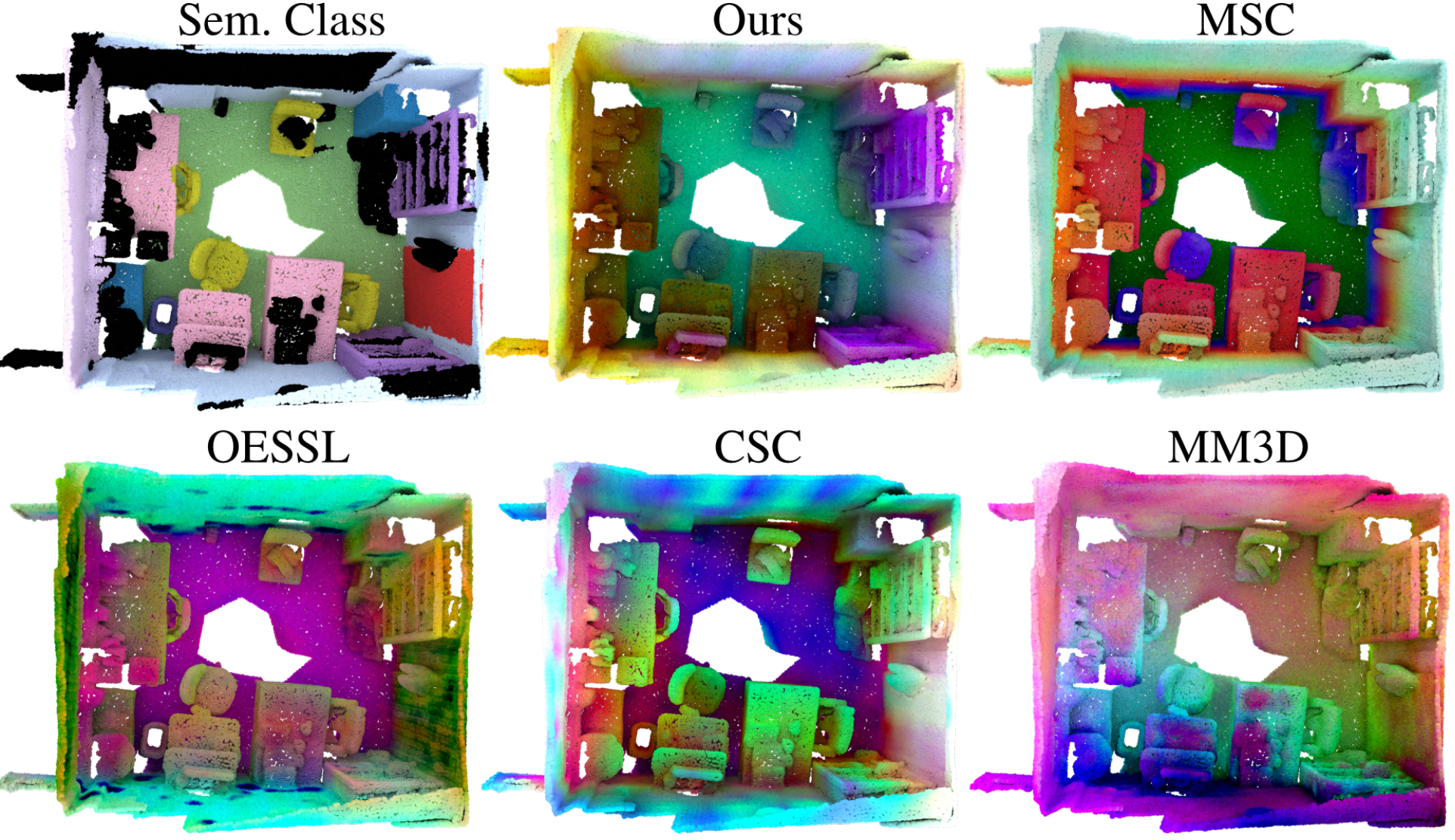}
    \caption{\textbf{Qualitative results.}
    Feature visualization of off-the-shelf features of our method and the baselines.
    Our learned features align with semantic classes better than existing methods.}
    \label{fig:qualitative}
\end{figure}

\subsection{Qualitative evaluation}
Following \cite{Ranzinger_2024_CVPR}, we use PCA to reduce the point features to three dimensions and visualize them as point colors.
Fig.~\ref{fig:qualitative} presents this visualization for all baselines compared to our model, where our learned features align with semantic classes better than existing methods.

%% file: 6_conclusions.tex
\section{Conclusions}
\label{sec:conclusions}

In this paper, we have introduced an evaluation protocol for self-supervised models tailored to 3D scenes that better reflects the capabilities of the representations learned by these models.
Moreover, we have introduced the first self-supervised model for 3D scene understanding that shows task-agnostic features capable of achieving supervised-like performance on several downstream tasks.
Our model not only outperforms all 3D self-supervised models tested, but also achieves better performance than 2D foundation models tasked to solve 3D problems, underlying the need for further 3D-native self-supervised representation learning approaches.
In the future, we would like to overcome the main limitation of our method, the reduced amount of data used for training, by consolidating a large dataset.

%% file: X_suppl.tex
\clearpage
\maketitlesupplementary
\appendix

{
    \hypersetup{linkcolor=black}
    \startcontents[sections]
    \printcontents[sections]{l}{1}{\setcounter{tocdepth}{2}}
}

\begin{figure*}
    \centering
    \includegraphics[width=\linewidth]{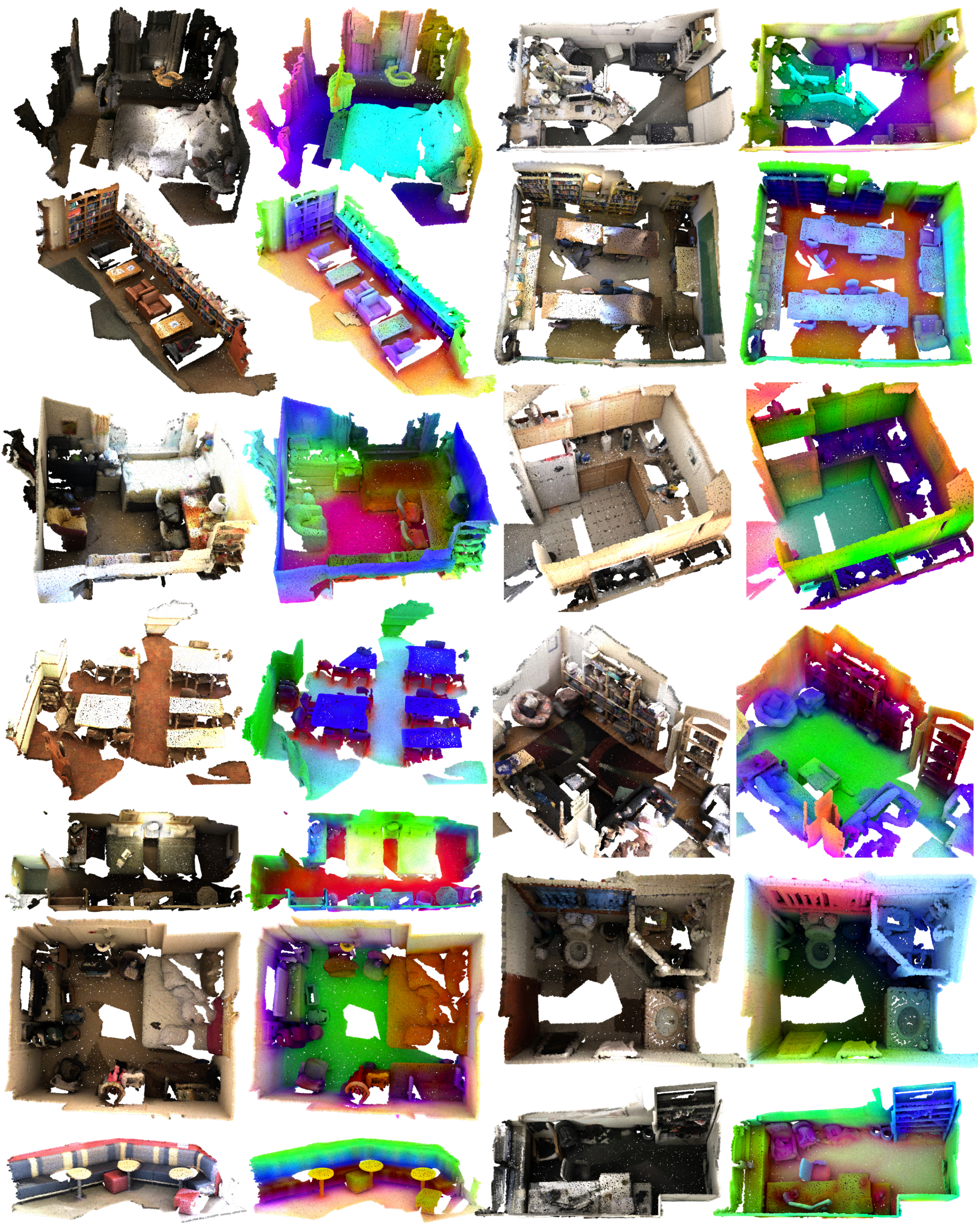}
    \caption{\textbf{Qualitative results.}
    Feature visualization of off-the-shelf features of our method and the baselines.
    Our learned features align with semantic classes better than existing methods.}
    \label{fig:qualitative_supp}
\end{figure*}

\section{Additional Qualitative Results}
\label{sec-sup:qualitative}

Fig.~\ref{fig:qualitative_supp} presents additional feature visualization of our self-supervised model for different 3D scenes.
We follow \cite{Ranzinger_2024_CVPR} and use PCA to reduce the point features to three dimensions and visualize them as point colors.
Results show that semantically similar objects result in similar features for all scenes.

\section{Additional Experiments}
\label{sec-sup:fine-tuning}

%%%%%%%%%%%%%%%%%%%%%%%%%%%%%%%%% FINE-TUNING
\subsection{Fine-Tuning}

Although not the main focus of this work, we also present results where our self-supervised model is used as a weight initialization step for fine-tuning on the downstream task.
In Tbl.~\ref{tbl:fine-tuning}, we present results on the semantic segmentation task on the three datasets used in our main experiments.
Our self-supervised model provides a significant improvement over supervised models trained from scratch and outperforms all existing self-supervised models.

\begin{table}[t]
\centering
\small
\setlength{\tabcolsep}{11pt}
\begin{tabular}{llccc}
    \toprule
      & & \multicolumn{1}{c}{ScN} & \multicolumn{1}{c}{ScN200} & \multicolumn{1}{c}{S3DIS} \\
      
    \midrule
    \multirow{8}{*}{\rotatebox[origin=c]{90}{Fine Tuning}} & \cbg \textcolor{SupColor}{SR-UNet} & \cbg \textcolor{SupColor}{72.2} & \cbg \textcolor{SupColor}{25.0} & \cbg \textcolor{SupColor}{68.2} \\
    & \hspace{2.5 mm}+ PC~\cite{xie_pointcontrast_2020} & 74.1 & 26.2 & 70.3 \\
    & \hspace{2.5 mm}+ CSC~\cite{hou_csc_2021} & 73.8 & 26.4 & 72.2 \\
    & \hspace{2.5 mm}+ MSC~\cite{wu_masked_2023} & 75.3 & 28.8 & -- \\
    & \hspace{2.5 mm}+ GC~\cite{wang_groupcontrast_2024} & 75.7 & 30.0 & 72.0 \\
    \cmidrule{2-5}
    & \cbg \textcolor{SupColor}{HUNet} & \cbg \textcolor{SupColor}{77.0} & \cbg \textcolor{SupColor}{35.4} & \cbg \textcolor{SupColor}{71.3} \\
    & \hspace{2.5 mm}+ MSC~\cite{wu_masked_2023} & 78.2 & 34.9 & 72.1 \\
    & \hspace{2.5 mm}+ Ours & \textbf{78.5} & \textbf{35.7} & \textbf{73.2} \\
   
    \bottomrule
\end{tabular}
\caption{\textbf{Fine-tuning.} Performance of different pre-trained methods after fine-tuning on the semantic segmentation task.}
\label{tbl:fine-tuning}
\end{table}

%%%%%%%%%%%%%%%%%%%%%%%%%%%%%%%%% OBJECT-CENTRIC

\subsection{Object-Centric Self-Supervised Methods}
\label{sec-sup:objcet-level}

Another important line of research focuses on self-supervised models pre-trained specifically on object-centric datasets.
While these models present strong performance in object-centric tasks, such as shape classification or shape segmentation, those models are not well suited for dense predictions usually required in 3D scene understanding, such as semantic segmentation of large indoor scenes.
However, due to the nature of these object-centric models, they are usually also evaluated on the 3D scene understanding task of object detection, where models need to predict the bounding box of objects instead of dense per-point instance segmentation maps.
Therefore, we use our self-supervised model as the 3D backbone in an object detection framework to compare our model with such methods.

\paragraph{Dataset.}
In this experiment, we use the ScanNet dataset~\cite{dai2017scannet}, and we report \ac{mAP} with \ac{IoU} thresholds of $0.5$ and $0.25$.
We use our model as the 3D backbone of the 3DETR~\cite{misra2021-3detr} object detection framework, and we evaluate our self-supervised model with two different protocols.
First, we obtain off-the-shelf features by freezing the 3D backbone while we train the remaining components of the 3DETR~\cite{misra2021-3detr} framework using our general-purpose features as input.
In the second protocol, we also fine-tune all the parameters of the 3D backbone using our self-supervised model as weight initialization.

\paragraph{Baselines.}
We compare our model to several state-of-the-art self-supervised models pre-trained on object-centric datasets and then fine-tuned on the object detection task.
These object-centric models use transformer-based architectures trained with different \ac{MIM} objectives.
While Point-Bert~\cite{Yu_2022_CVPR}, Point-MAE~\cite{pang2022PointMAE}, and MaskPoint~\cite{liu2022masked} use a non-hierarchical architecture, Point-M2AE~\cite{zhang2022point} use a hierarchical model with a bottom-up masking approach.
However, all models reconstruct the point coordinates from the last layer in the model.

\begin{table}[t]
\centering
\small
\setlength{\tabcolsep}{14pt}
\begin{tabular}{lcc}
    \toprule
      & \multicolumn{1}{c}{mAP@25} & \multicolumn{1}{c}{mAP@50}\\
      
    \midrule
    \cbg \textcolor{SupColor}{3DETR~\cite{misra2021-3detr}} & \cbg \textcolor{SupColor}{62.1} & \cbg \textcolor{SupColor}{37.9} \\
     \hspace{2.5 mm}+ Point-Bert~\cite{Yu_2022_CVPR} & 61.0 & 38.3 \\
     \hspace{2.5 mm}+ Point-MAE~\cite{pang2022PointMAE} & 63.4 & 40.6 \\
     \hspace{2.5 mm}+ MaskPoint~\cite{liu2022masked} & 63.4 & 40.6 \\
     \hspace{2.5 mm}+ Point-M2AE~\cite{zhang2022point} & 66.3 & 48.3 \\
    \midrule
     \hspace{2.5 mm}+ Ours (Lin.) & 65.6 & 40.2 \\
     \hspace{2.5 mm}+ Ours (FT) & \textbf{71.3} & \textbf{52.2} \\
   
    \bottomrule
\end{tabular}
\caption{\textbf{Object detection.} Comparison of our off-the-shelf features to fine-tuning object-centric self-supervised methods.}
\label{tbl:obj-centric}
\end{table}

\paragraph{Results.}
Tbl.~\ref{tbl:obj-centric} presents the results of our experiments.
Our off-the-shelf features, \emph{Lin.} on Tbl.~\ref{tbl:obj-centric}, present a competitive performance, outperforming most existing object-centric self-supervised methods.
When we further fine-tune our model on the downstream task, \emph{FT} on Tbl.~\ref{tbl:obj-centric}, we outperform all models by a large margin.
These results are in line with the results presented by Xie~\etal~\cite{xie_pointcontrast_2020} and highlight the need for scene-centric self-supervised methods.

\begin{table}
\centering
\small
\setlength{\tabcolsep}{7pt}
\begin{tabular}{lcccc}
    \toprule
      & \multicolumn{2}{c}{Obj. Det.} & \multicolumn{2}{c}{Sem. Seg.} \\
      \cmidrule(l{2pt}r{2pt}){2-3}
      \cmidrule(l{2pt}r{2pt}){4-5}
      & \multicolumn{1}{c}{mAP@25} & \multicolumn{1}{c}{mAP@50} & \multicolumn{1}{c}{ScN} & \multicolumn{1}{c}{S3DIS}\\
      
    \midrule
     Bridge3D~\cite{chen2023bridging} & 65.3 & 44.2 & 73.9 & 70.2 \\
     SAM-MAE~\cite{chen2024samguided} & 68.2 & 48.4 & 75.4 & 71.8 \\
    \midrule
     Ours (Lin.) & 65.6 & 40.2 & 68.7 & 59.5 \\
     Ours (FT) & \textbf{71.3} & \textbf{52.2} & \textbf{78.5} & \textbf{73.2} \\
   
    \bottomrule
\end{tabular}
\caption{\textbf{2D-3D KD.} Comparison to methods that rely on knowledge distillation from 2D foundation models.}
\label{tbl:2d-3d}
\end{table}

%%%%%%%%%%%%%%%%%%%%%%%%%%%%%%%%% 2D-3D KD

\subsection{2D-3D Knowledge Distillation Methods}
\label{sec-sup:2d-3d}

Since general models for 3D scene understanding are not available, recent works have proposed distilling knowledge from 2D foundation models.
While Bridge3D~\cite{chen2023bridging} combines several 2D foundation models for knowledge distillation into a non-hierarchical 3D transformer architecture, SAM-MAE~\cite{chen2024samguided} uses SAM~\cite{kirillov2023segment} to mask objects in 3D space and a \ac{MIM} objective to train the same model architecture.
We compare our self-supervised model to these models fine-tuned on object detection and semantic segmentation tasks.

\paragraph{Result.}
Tbl.~\ref{tbl:2d-3d} presents the results of this experiment.
While our linear probing setup is not able to achieve the same performance as the baselines, when fine-tuned, our model can outperform them in all experiments.

\begin{table*}[htb]
\centering
\begin{subtable}{0.24\linewidth}
\centering
\setlength{\tabcolsep}{16pt}
\begin{tabular}{cc}
    \toprule
    No Mask & 50.7 \\
    \cbg Mask& \cbg \textbf{66.8} \\
    \bottomrule
\end{tabular}
\caption{\textbf{Masking.} Patch supervision with vs without masking.}
\end{subtable}
\hfill
\begin{subtable}{0.24\linewidth}
\centering
\setlength{\tabcolsep}{20pt}
\begin{tabular}{cc}
    \toprule
    Last & 60.5 \\
    \cbg All & \cbg \textbf{66.8} \\ 
    \bottomrule
\end{tabular}
\caption{\textbf{Supervision.} Layers in the hierarchy used in the loss.}
\end{subtable}
\hfill
\begin{subtable}{0.24\linewidth}
\centering
\setlength{\tabcolsep}{14pt}
\begin{tabular}{cc}
    \toprule
    top-down & 62.4 \\
    \cbg bottom-up & \cbg \textbf{66.8} \\
    \bottomrule
\end{tabular}
\caption{\textbf{Mask strategy.} Masking hierarchy top-down vs bottom-up.}
\end{subtable}
\hfill
\begin{subtable}{0.24\linewidth}
\centering
\setlength{\tabcolsep}{10pt}
\begin{tabular}{cc}
    \toprule
    SparseConv & 61.8 \\
    MHA & 52.3 \\
    \cbg HUNet & \cbg \textbf{66.8} \\     
    \bottomrule
\end{tabular}
\caption{\textbf{Model.} Types of model used.}
\end{subtable}
\caption{\textbf{Ablation studies.} Evaluation of the different components of our framework on the task of semantic segmentation on ScanNet.}
\label{tbl:abl}
\end{table*}

%%%%%%%%%%%%%%%%%%%%%%%%%%%%%%%%% ABLATIONS

\section{Ablation Studies}
\label{sec-sup:ablations}

In this section, we describe the ablation studies conducted to validate our design choices.
For all our experiments, we report linear probing performance on the task of semantic segmentation on ScanNet.
Unless otherwise stated, due to the large training times of the self-supervise stage, we perform our ablation studies on a smaller model that takes as input a coarser voxelization of the scene, $4\,$cm voxels, and we train our models for $800$ epochs instead of $1800$.
For more details of the experimental setup and model used, we refer the reader to Sec.~\ref{sec-sup:det-exp-setup}.

\subsection{Masking}
In this experiment, we evaluate the importance of our \emph{Masked Scene Modeling} objective.
We train a model with our full framework and the same model without our masking strategy.
In this version of our framework, the crops given to the student model are not masked, and the full crop is processed by the model.
Then, the training objective is the prediction of deep features from the teacher model, which has access to a full view of the scene with different data augmentation.
This objective is similar to the self-distillation objective used in MM3D~\cite{Xu_2023_CVPR}.
Tbl.~\ref{tbl:abl} (a) presents the results of this experiment.
We can see that the proposed \emph{Masked Scene Modeling} objective is essential for learning semantically relevant features, leading to an improvement of more than \textbf{+16} points.

\subsection{Hierarchical Supervision}
In this experiment, we measure the importance of the hierarchical reconstruction objective.
We compare our full framework with a model trained with supervision only on the last layer of the decoder, a common practice in existing self-supervised approaches for 3D scenes~\cite{hou_csc_2021,wu_masked_2023,xie_pointcontrast_2020}.
Tbl.~\ref{tbl:abl} (b) shows that supervising only the last layer leads to a gap in performance of more than \textbf{+6}.
This experiment aligns with the findings of our pilot study and highlights the importance of hierarchical supervision when training hierarchical architectures.

\subsection{Masking Strategy}
We also compare our bottom-up masking strategy with a traditional top-down approach, similar to the one used in MSC~\cite{wu_masked_2023}.
In this approach, instead of incorporating the masked patches in the decoder, we add them in the encoder with the corresponding learnable token.
We can see that in Tbl.~\ref{tbl:abl} (c), even though a top-down approach can lead to relatively good features, our bottom-up approach leads to semantically richer features with more than \textbf{+4} points of improvement on the downstream task.

\subsection{Model Architecture}
We also evaluate the effect of the model architecture used.
We trained two additional models, one only based on Sparse convolutions without \ac{MHA} blocks, and another one with \ac{MHA} instead of ResNet blocks as in Ptv3~\cite{wu2024ptv3}.
Tbl.~\ref{tbl:abl} (d) indicates that the model using only sparse convolutions provides lower performance than our hybrid architecture.
Moreover, the model with only \ac{MHA} layers significantly reduces the performance on the downstream task.
This is due to the additional constraints of such models, where a lower learning rate is necessary to avoid unstable training.
Although we believe that an exhaustive hyperparameter search could lead to an improvement of such models, our hybrid model architecture is robust to higher learning rates and, therefore, easier to train.

\begin{figure}
    \centering
    \includegraphics[width=\linewidth]{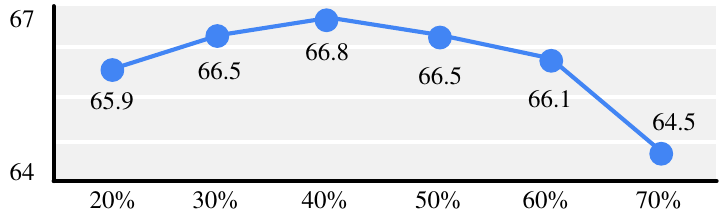}
    \caption{\textbf{Masking ratio.}
    Linear probing performance for different masking ratios.}
    \label{fig:masking}
\end{figure}

\subsection{Masking Ratio}
Additionally, we measure the influence of the masking ratio on the final performance of the model.
We evaluated a range of ratios from $20\,$\% to $70\,$\% with intervals of $10\,$\% and plot the results in Fig.~\ref{fig:masking}.
The results show that the framework is relatively robust to the masking ratio used, achieving similar performance for ratios between $30\,$\% and $60\,$\%, with the highest value obtained at $40\,$\%.
However, smaller ratios, such as $20\,$\%, or too high, such as $70\,$\%, lead to a significant drop in performance.

\begin{table}
    \centering
    \setlength{\tabcolsep}{26pt}
    \begin{tabular}{lcc}
        \toprule
       Layer & Alone & Remove \\
        \midrule
        1 & 28.3 & 67.1 \\
        2 & 43.5 & 67.0 \\
        3 & 54.8 & 67.0 \\
        4 & 62.8 & 66.6 \\
        5 & 62.4 & 64.4 \\
        \midrule
        All & \multicolumn{2}{c}{68.7} \\
        \bottomrule
    \end{tabular}
    \caption{\textbf{Layer importance.} Comparison of the performance on the linear probing setup when only one layer is used (\textit{Alone}) or when all layers except one are used (\textit{Remove}).}
    \label{tab:hierarchical-layers}
\end{table}

\subsection{Layer Importance}
To expand our pilot study, we further evaluate the importance of the different layers on the performance of our final model.
First, we evaluate the linear probing abilities when only one layer is used as input.
Then, we evaluate the effect of using all layers except one for the same linear probing setup.
Tbl.~\ref{tab:hierarchical-layers} present the results of this experiment.
Results show that, for all layers, using the output of one layer alone (\textit{Alone} in Tbl.~\ref{tab:hierarchical-layers}) leads to a lower performance than using a concatenation of all of them.
Moreover, results also show that using all layers except one (\textit{Remove} in Tbl.~\ref{tab:hierarchical-layers}) also leads to a degradation in performance in all cases.
This experiment shows the importance of all layers, indicating that each layer provides complementary information.

Additionally, we also evaluate different methods of combining such features.
We compare the concatenation of features used in all of our experiments ($68.7$ mIoU), to a setup where the features are aggregated with a sum operator ($68.7$ mIoU) and to a setup where the features are aggregated with a learned weighted sum ($68.8$ mIoU).
Our results show that there is no significant difference between these methods.

\begin{figure}
    \centering
    \includegraphics[width=\linewidth]{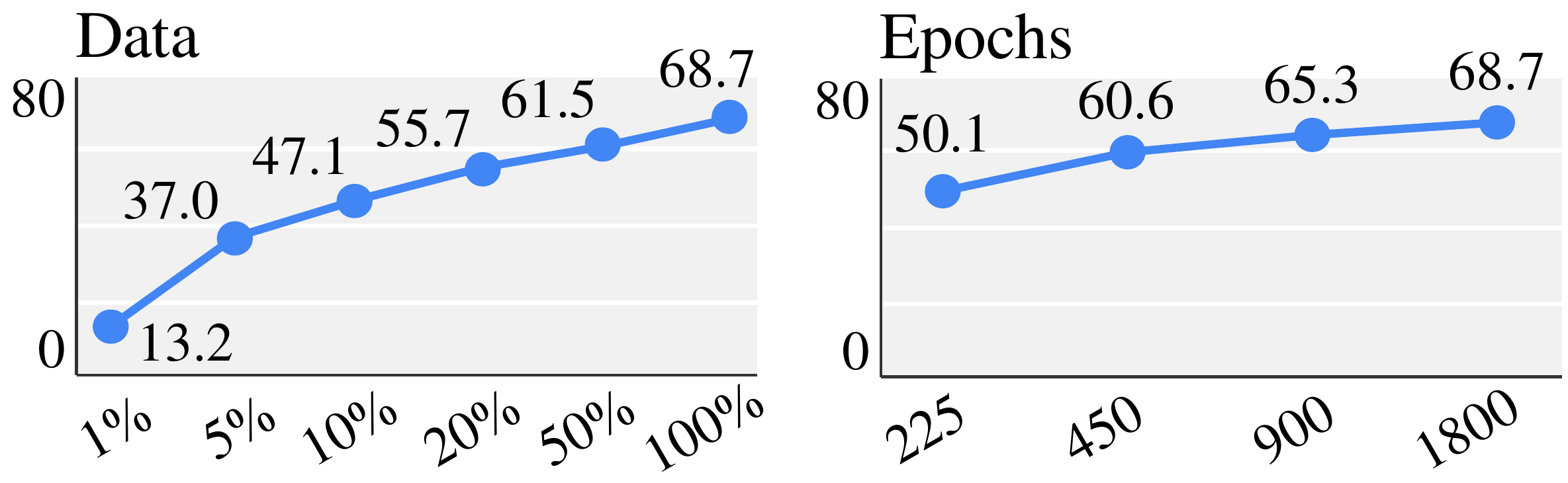}
    \caption{\textbf{Scalability experiments.}
    Evaluate the performance of the model under reduced data used for pre-training and reduced number of epochs.}
    \label{fig:scalability}
\end{figure}

\subsection{Scaling Properties}

Moreover, we evaluate the scaling abilities of our framework \wrt the data used for pre-training and the number of epochs.
For this setup, we use our full model and configuration as in the main experiments in the paper.
Fig.~\ref{fig:scalability} presents the results of these experiments.
Results show that more data and longer pre-training yield significant improvements for linear probing on semantic segmentation.
This highlights the importance of additional data and training in self-supervised objectives and paves the road for future improvements of our method.

\subsection{NN Robustness}

Lastly, we evaluate the robustness of the NN evaluation protocol \wrt the distance metric used to compare features.
We compare the L2 distance used in all our experiments ($65.7$ mIoU), to the L1 distance ($66.4$ mIoU) and to the cosine distance ($66.0$ mIoU).
Although other distance metrics yield slightly better performance, the experiment indicates that the evaluation protocol is robust to the distance metric chosen for evaluation.

\section{Detailed experimental setup}
\label{sec-sup:det-exp-setup}

\begin{table}
    \centering
    \setlength{\tabcolsep}{15pt}
    \begin{tabular}{lc}
        \toprule
        Config & Value \\
        \midrule
        Voxel size & 2\,cm \\
        Norm layers  & RMSNorm~\cite{zhang2019rmsnorm} \\
        Downsample & Strided SparseConv \\
        Upsample & Transpose SparseConv \\
        Serialization &  Z + TZ + H + TH \cite{wu2024ptv3}\\
        Block bias & False \\
        Att. drop & 0.1 \\
        Drop path & 0.4 \\
        Activation func. & GELU~\cite{hendrycks2016gaussian} \\
        FF layer & GEGLU~\cite{shazeer2020glu} \\
        FF ratio & 4 \\
        Enc. channels & [32, 64, 128, 256, 384] \\
        Enc. ResNet & [2, 2, 2, 2, 2] \\
        Enc. MHA & [0, 0, 0, 2, 2] \\
        Enc. MHA Window & [0, 0, 0, 1024, 1024] \\
        Enc. MHA \# Heads & [0, 0, 0, 32, 48] \\
        Dec. channels & [64, 96, 128, 256, 384] \\
        Dec. ResNet & [2, 2, 2, 2, 2] \\
        Dec. MHA & [0, 0, 0, 2, 2] \\
        Dec. MHA Window & [0, 0, 0, 1024, 1024] \\
        Enc. MHA \# Heads & [0, 0, 0, 32, 48] \\
        \bottomrule
    \end{tabular}
    \caption{\textbf{Model configuration.}}
    \label{tab:model_config}
\end{table}

\begin{table}
    \centering
    \setlength{\tabcolsep}{15pt}
    \begin{tabular}{lc}
        \toprule
        Config & Value \\
        \midrule
        Voxel size & 4\,cm \\
        Norm layers  & RMSNorm~\cite{zhang2019rmsnorm} \\
        Downsample & Strided SparseConv \\
        Upsample & Transpose SparseConv \\
        Serialization &  Z + TZ + H + TH \cite{wu2024ptv3}\\
        Block bias & False \\
        Att. drop & 0.1 \\
        Drop path & 0.4 \\
        Activation func. & GELU~\cite{hendrycks2016gaussian} \\
        FF layer & GEGLU~\cite{shazeer2020glu} \\
        FF Ratio & 4 \\
        Enc. channels & [64, 128, 256, 384] \\
        Enc. ResNet & [2, 2, 2, 2] \\
        Enc. MHA & [0, 0, 2, 2] \\
        Enc. MHA Window & [0, 0, 1024, 1024] \\
        Enc. MHA \# Heads & [0, 0, 32, 48] \\
        Dec. channels & [96, 128, 256, 384] \\
        Dec. ResNet & [2, 2, 2, 2] \\
        Dec. MHA & [0, 0, 2, 2] \\
        Dec. MHA Window & [0, 0, 1024, 1024] \\
        Enc. MHA \# Heads & [0, 0, 32, 48] \\
        \bottomrule
    \end{tabular}
    \caption{\textbf{Model configuration for ablation studies.}}
    \label{tab:model_config_abl}
\end{table}

\subsection{Model architecture}
We designed a Hybrid UNet architecture (HUnet) combining standard ResNet blocks~\cite{he2016deep} with serialization transformer layers as in PTv3~\cite{wu2024ptv3}.
However, contrary to PTv3~\cite{wu2024ptv3}, we use sliding-window attention as in LongFormer~\cite{beltagy2020longformer} since this eliminates the need for padding and makes the receptive field adaptive.
Moreover, we do not include xCPE~\cite{wu2024ptv3} in such layers since the ResNet blocks can act as conditional positional encoding.
Furthermore, following the design of Stable Diffusion~\cite{rombach2021highresolution}, we only included the \ac{MHA} layers in the lowest resolution levels of the model, making the model faster and more stable to different learning rates.
Tbl.~\ref{tab:model_config} presents a detailed description of the different components of our architecture, such as channels per level, number of layers per level, or activation function used.
We also provide the configuration of the model used for the ablation studies in Tbl.~\ref{tab:model_config_abl}.
For these experiments, we used a smaller model with one level less in the encoder and decoder, which takes bigger voxels of $4\,$cm as input.

\subsection{Experiment hyperparameters}

\paragraph{Self-supervised training.}
We build our self-supervised framework on top of the codebase Pointcept~\cite{pointcept2023}.
The hyperparameters used for training our self-supervised model are described in Tbl.~\ref{tab:self_sup_config}.
As data augmentation, we use the default augmentations for indoor semantic segmentation of PTv3~\cite{wu2024ptv3}.
We only increase the number of points per crop as described in Tbl.~\ref{tab:self_sup_config}.

\begin{table}
    \centering
    \setlength{\tabcolsep}{28pt}
    \begin{tabular}{lc}
        \toprule
        Config & Value \\
        \midrule
        Optimizer & AdamW~\cite{loshchilov2018decoupled} \\
        Betas & (0.9, 0.95) \\
        Weight decay & 0.05 \\
        Learning rate & 0.0015 \\
        LR Scheduler & Cosine \\
        Batch size & 12 \\
        Epochs & 1800\\
        Warmup epochs & 60\\
        Crop size & 240000 \\
        Mask Ratio & 0.4 \\
        Teacher mom. & 0.996 $\rightarrow$ 1.0 \\
        \bottomrule
    \end{tabular}
    \caption{\textbf{Self-supervised training configuration.}}
    \label{tab:self_sup_config}
\end{table}

\paragraph{Linear probing - Semantic and Instance segmentation.}
We use the codebase Pointcept~\cite{pointcept2023} for our linear probing experiments in the downstream tasks of semantic and instance segmentation.
The hyperparameters used in these experiments are described in Tbl.~\ref{tab:semseg_config} and Tbl.~\ref{tab:inst_config}.
For data augmentation, we use the default configuration of PTv3~\cite{wu2024ptv3}.

\begin{table}
    \centering
    \setlength{\tabcolsep}{6pt}
    \begin{tabular}{lccc}
        \toprule
        Config & \multicolumn{3}{c}{Value} \\
        \cmidrule{2-4}
        & ScanNet & ScanNet200 & S3DIS\\
        \midrule
        Optimizer & \multicolumn{3}{c}{AdamW~\cite{loshchilov2018decoupled}} \\
        Betas & \multicolumn{3}{c}{(0.9, 0.95)} \\
        Weight decay & \multicolumn{3}{c}{0.01} \\
        Learning rate & \multicolumn{3}{c}{0.01} \\
        LR Scheduler & \multicolumn{3}{c}{Cosine} \\
        Batch size & \multicolumn{3}{c}{8} \\
        Epochs & 200 & 200 & 100\\
        Warmup epochs & 2 & 2 & 1\\
        Crop size & 120000 & 120000 & 200000  \\
        \bottomrule
    \end{tabular}
    \caption{\textbf{Linear probing config. for semantic segmentation.}}
    \label{tab:semseg_config}
\end{table}

\begin{table}
    \centering
    \setlength{\tabcolsep}{6pt}
    \begin{tabular}{lccc}
        \toprule
        Config & \multicolumn{3}{c}{Value} \\
        \cmidrule{2-4}
        & ScanNet & ScanNet200 & S3DIS\\
        \midrule
        Optimizer & \multicolumn{3}{c}{SGD} \\
        Momentum & \multicolumn{3}{c}{0.9} \\
        Weight decay & \multicolumn{3}{c}{0.0001} \\
        Learning rate & \multicolumn{3}{c}{0.1} \\
        LR Scheduler & \multicolumn{3}{c}{PolyR} \\
        Batch size & 12 & 12 & 8 \\
        Epochs & 200 & 200 & 100\\
        Crop size & -- & -- & 200000  \\
        \bottomrule
    \end{tabular}
    \caption{\textbf{Linear probing config. for instance segmentation.}}
    \label{tab:inst_config}
\end{table}

\paragraph{Coss-Attention - Visual grounding.}
Given a 3D point cloud with associated features, 3D ground truth bounding boxes of objects, and a text description, the model is tasked to select the object that matches the text description.
We encode the text with the CLIP text encoder~\cite{radford2021learning} and use the attention head of Zhand~\etal~\cite{zhang2023multi3drefer} composed of self- and cross-attention layers.
The cross-attention layers combine the text CLIP embeddings and object features (obtained from aggregating point features inside object bounding boxes).
The output of the model is a probability per object.
Then, we train the model using cross-entropy loss, since the task can be formulated as a classification problem where the object matching the text description should have the highest probability.
We use the codebase of Multi3DRefer~\cite{zhang2023multi3drefer} and the hyperparameters used in these experiments are described in Tbl.~\ref{tab:visual_grounding_config}.
For data augmentation, we use the default configuration of PTv3~\cite{wu2024ptv3} for the task of instance segmentation.

\begin{table}
    \centering
    \setlength{\tabcolsep}{25pt}
    \begin{tabular}{lc}
        \toprule
        Config & Value \\
        \midrule
        Optimizer & AdamW~\cite{loshchilov2018decoupled} \\
        Betas & (0.9, 0.95) \\
        Weight decay & 0.00001 \\
        Learning rate & 0.0005 \\
        LR Scheduler & Cosine \\
        Batch size & 12 \\
        Epochs & 10\\
        Warmup epochs & 1\\
        \bottomrule
    \end{tabular}
    \caption{\textbf{Visual grounding configuration.}}
    \label{tab:visual_grounding_config}
\end{table}

\paragraph{Object detection.}
In these experiments, we use the object detection framework 3DETR~\cite{misra2021-3detr}.
For the linear probing and fine-tuning experiments, we use the same configuration described in Tbl.~\ref{tab:object_det_config}.
For data augmentation, we use the default configuration of 3DETR~\cite{misra2021-3detr}.

\begin{table}
    \centering
    \setlength{\tabcolsep}{25pt}
    \begin{tabular}{lc}
        \toprule
        Config & Value \\
        \midrule
        Optimizer & AdamW~\cite{loshchilov2018decoupled} \\
        Betas & (0.9, 0.95) \\
        Weight decay & 0.1 \\
        Learning rate & 1e-6 \\
        LR Scheduler & Cosine \\
        Batch size & 24 \\
        Epochs & 1080\\
        Warmup epochs & 9\\
        Clip gradients & 0.1 \\
        \# queries & 256 \\
        \# points & 2048 \\
        \bottomrule
    \end{tabular}
    \caption{\textbf{Object detection configuration.}}
    \label{tab:object_det_config}
\end{table}

\paragraph{Fine-tuning - Semantic segmentation.}
For fine-tuning on the task of semantic segmentation, we use a different configuration than the one used in our linear probing experiments.
The hyperparameters of these experiments are described in Tbl.~\ref{tab:fine_tuning_config}.

\begin{table}
    \centering
    \setlength{\tabcolsep}{6pt}
    \begin{tabular}{lccc}
        \toprule
        Config & \multicolumn{3}{c}{Value} \\
        \cmidrule{2-4}
        & ScanNet & ScanNet200 & S3DIS\\
        \midrule
        Optimizer & \multicolumn{3}{c}{AdamW~\cite{loshchilov2018decoupled}} \\
        Betas & \multicolumn{3}{c}{(0.9, 0.95)} \\
        Weight decay & \multicolumn{3}{c}{0.05}  \\
        Learning rate & \multicolumn{3}{c}{0.001} \\
        LR Scheduler & \multicolumn{3}{c}{Cosine} \\
        Batch size & 48 & 48 & 32 \\
        Epochs & 200 & 200 & 500\\
        Warmup epochs & 2 & 2 & 20\\
        Crop size & 120000 & 120000 & 100000  \\
        \bottomrule
    \end{tabular}
    \caption{\textbf{Fine-tuning config. for semantic segmentation.}}
    \label{tab:fine_tuning_config}
\end{table}

\paragraph{Masked Scene Context.}
For training our model with the baseline MSC~\cite{wu_masked_2023}, we use different hyperparameters than the ones recommended by the authors.
Our model trained with the default parameters leads to subpar performance, obtaining less than $20$ mIoU on the task of linear probing for semantic segmentation on ScanNet.
Therefore, we modified the number of training epochs to $1800$ instead of $600$ and the optimizer from SGD to AdamW~\cite{loshchilov2018decoupled}. 
These small changes lead to an increase in performance, as reported in the main experiments of this paper.